\RequirePackage{fix-cm}
\documentclass[twocolumn]{svjour3}          
\smartqed  
\usepackage{graphicx}
\usepackage{amsmath}
\usepackage{enumerate}
\usepackage{float} 
\usepackage{subfigure}
\usepackage{mathrsfs}
\usepackage{color}
\journalname{The Visual Computer}
\begin{document}

\title{Semantic-Aware Label Placement for Augmented Reality in Street View
}


\author{Jianqing Jia\and
Semir Elezovikj\and
Heng Fan\and
Shuojin Yang\and
Jing Liu\and
Wei Guo\and
Chiu C. Tan\and         
Haibin Ling
}


\institute{Jianqing Jia, Shuojin Yang, Jing Liu, Wei Guo \at
              Key Laboratory of Augmented Reality. College of Mathematics and Information Science, Hebei Normal University,  Shijiazhuang 050024, Hebei, China \\
           \and
           Semir Elezovikj, Heng Fan, Chiu C. Tan, Haibin Ling \at
              Department of Computer and Information Sciences, Temple University,  1925 North 12th St.,Philadelphia, PA 19122, USA
}

\date{Received: date / Accepted: date}

\maketitle

\begin{abstract}
In an augmented reality (AR) application, placing labels in a manner that is clear and readable without occluding the critical information from the real-world can be a challenging problem. This paper introduces a label placement technique for AR used in street view scenarios. We propose a semantic-aware task-specific label placement method by identifying potentially important image regions through a novel feature map, which we refer to as \textit{guidance map}. Given an input image, its saliency information, semantic information and the task-specific importance prior are integrated in the guidance map for our labeling task. To learn the task prior, we created a label placement dataset with the users' labeling preferences, as well as use it for evaluation. Our solution encodes the constraints for placing labels in an optimization problem to obtain the final label layout, and the labels will be placed in appropriate positions to reduce the chances of overlaying important real-world objects in street view AR scenarios. The experimental validation shows clearly the benefits of our method over previous solutions in the AR street view navigation and similar applications.
\end{abstract}

\section{Introduction}
\label{intro}
Augmented Reality (AR) technology enhances the physical world with digital information by overlaying visual information such as in the form of labels \cite{ref1}. Well-placed informative labels can provide accurate instructions for important objects such as landmarks, road signs, and so on. However, the task of placing labels in the user’s view can be challenging, especially for dense scenes. The label annotations should be placed to avoid overlapping and occluding real-world textual objects and scene features. The readability of the text annotations themselves depends strongly on the background color and texture \cite{ref2,ref3}. In the area of dynamic computer generated label layouts, labels should satisfy readability, unambiguity, aesthetics and frame coherence, especially in AR scenes \cite{ref4} where these considerations could be more challenging.

In common AR applications, the computer-generated labels are usually registered based on the object’s geographical locations, which are usually given as points of interest (POI) with their respective GPS position. The precise models of the environment are difficult to obtain, especially since the models can include dynamic objects, for the following reasons. First,  such a model  usually cannot capture fine details that determine the background clutter in images. Second, the general model may not include all moving objects or able track them. For example, a driver ignoring a pedestrian may cause a traffic accident. Finally, inaccurate registration of sensor-based tracking may not lead to productive use of the additional scene knowledge. To obtain appropriate layout in all situations, an image-driven label placement method for AR is required.

Prior work on image-based label placement algorithms generally used some visual saliency map to highlight prominent regions in an image that attracts human attention. However, these methods often suffer from the following problems: 
\begin{itemize}
    \item They ignore the semantic information in the user’s view, which can be very useful in enhancing the understanding of human interest regions
    \item Previous works are task-unaware, and make little use of the user's understanding of the specific scene, interests and preferences. The saliency algorithm by itself has limitations with regards to label placement. Specifically, the aim of image saliency detection is to highlight visually salient regions in an image, whereas the most salient region may not always correspond to the most important region for a specific scenario such as during driving. 
    \item To the best of our knowledge, there are no standard benchmarks for quantitatively evaluate label placement methods. 
    \item Previously used saliency models are typically outdated, e.g., the saliency detection method proposed by Achanta \cite{ref36} is mostly used in label placement area, leaving unexplored many recent advances \cite{WangLFSL19arXiv}in saliency analysis.
\end{itemize}

Keeping these limitations in mind, we propose an image-driven semantic-aware label placement approach to obtain an appropriate layout, which can improve the visual quality of annotated AR environments in street view scenarios. Unlike prior works which mostly use the saliency map to highlight important regions to aid them with the label layout task, we introduce a new feature map called \textbf{guidance map} to characterize the important regions, which can be regarded as a task-specific importance map. The guidance map consists of three parts corresponding respectively to the saliency information, the semantic information, and the task-specific importance prior which is automatically learned from the dataset. For the saliency information, we use a state-of-the-art deep-learning based saliency detection method, instead of previous methods using traditional saliency detection algorithm like IG \cite{ref36}. For the semantic information, we use the deep-learning based semantic segmentation algorithm proposed by Chen \cite{ref51}. For the task-specific importance prior, we automatically learn the prior using the manually labeled dataset. We created a manual label placement dataset to provide the task-specific adjustment to our system as the importance prior. The dataset also serves as quantitative evaluation benchmark. Finally, the proposed evaluation metrics brings a convincing quantitative comparison with the state-of-the-art label placement methods.

In this work, we only consider external labeling and do not consider internal labeling. Our labels will be placed in such a way that interference with task specific important real world information will be reduced. Although we mainly focus on street view in this paper, our task-specific label placement framework is not restricted to providing special benefits only to applications like AR street view navigation, but can be also applied to many other AR applications which are lacking scene knowledge without loss of generality.\par

This paper makes the following contributions:
\begin{enumerate}
    \item We introduce a new feature map called Guidance Map. In addition to using saliency information of the image, we add the semantic information and task-specific importance prior into the label placement optimization, to make the Guidance Map more aligned to the task specific important regions. 
    \item We manually collect a label placement dataset. Different from former task-unaware saliency detection, we use the dataset to learn task-specific importance priors. Morever, it serves as a quantitative evaluation benchmark.
    \item We integrate the latest state-of-the-art saliency model to further improve the label placement performance.
    \item We define the evaluation metrics in the experiment to quantitatively evaluate the label placement results. 
\end{enumerate}
\par

The rest of the paper is organized as follows. In Section 2, we cover related work in the label placement field, especially image-based layout techniques. In Section 3, we introduce our semantic-aware task-specific label placement method and details of data collection. Experimental results in Section 4 show the efficiencies of the proposed method. Finally, we conclude the paper and describe future work in Section 5.

\section{Related Work}
\label{sec:relat}
Annotation placement has been discussed in augmented reality view management (ARVM) \cite{ref27} that aims to determine positions and rendering styles of visual annotations based on the camera poses and the 3D scene information for augmented views. Previous ARVM approaches can be grouped into two categories: geometry-based ones and image-based ones. 

In the geometry-based approaches, camera poses and 3D scene information are used to determine the annotation placement. Bell et al. \cite{ref6} proposed to manage the view in interactive 3D user interfaces by solving the point-feature annotation placement problem using greedy algorithms. Cmolik and Bittner \cite{ref8} place the annotations for 3D objects by combining greedy algorithms with fuzzy logic. Tenmoku et al. \cite{ref9} propose a view management method to emphasize objects viewed by end users, leveraging 3D models of the real scene to locate annotations. Makita et al. \cite{ref10} overlay annotations of objects such as pedestrians for wearable AR, making use of the positions and shapes of objects to compute the positions of annotations via minimization. Zhang et al. \cite{ref11} propose an annotating system for augmenting tourist videos by registering videos to 3D models and the label placement is achieved using a dynamic programming algorithm.

More recent work by Iwai et al. \cite{ref12} proposed an annotation arrangement method for projection-based AR applications. Each annotation is superimposed directly onto a nonplanar and textured surface based on positions by minimizing an energy function using a genetic algorithm. Tatzgern et al. \cite{ref13} cluster similar items and select a single representation for each cluster to reduce the annotation complexity of crowded objects. In a different work, Tatzgern et al. \cite{ref14} apply 3D geometric constraints to manage the annotation placement in the 3D object space and attain temporal coherence. Kishishita et al. perform experiments on two view management methods for a wide field of view augmented reality display. 

Most prior works under the geometry-based approach formulate the view management problem as an optimization problem and propose different algorithms to obtain annotation positions in each frame. Some researchers focus on applying geometry-based approaches to display visual information for head-mounted display (HMD). For example, Lauber and Butz \cite{ref19} propose a layout management technique for HMD in cars. The annotations are rearranged based on the driver’s head rotations to avoid the label occlusions in the driver’s view. Orlosky et al. \cite{ref52} evaluate user tendencies through a see-through HMD to improve view management algorithms. A user tendency is found to arrange the label locations near the center of the viewing field.

A number of image-driven attempts have been made for improving label placement in augmented reality. In these methods, a weighted linear combination of factors that affect visibility is often utilized \cite{ref7,ref10}, however, the method of finding optimum weights has not been shown in detail. Leykin et al. \cite{ref15} used a pattern recognition method to automatically obtain the readability of annotations over textured backgrounds. However, the method fails to include an explicit algorithm to choose the appropriate position for the labels. Rosten et al. \cite{ref16} identify unimportant regions of an image using FAST features. The positions of annotations are decided by considering the occluded image features when the labels are placed in particular positions. However, the technique is demonstrated with a only few labels and for indoor scenes. Tanaka et al. \cite{ref17} proposed a color-based viewability estimation where they use the averages of RGB components, the S component in HSV color space, and the Y component in YCbCr color space. Grasset et al. \cite{ref18} proposed a framework that combines visual saliency with edge analysis to identify image regions suitable for label placement. They used the algorithm proposed by Achanta et al. \cite{ref36} for computing the saliency maps. We think that the labeling results can be improved, especially for scenes like automated driving. Fujinami et al. \cite{ref56,ref57} proposed a view management method for spatial AR called VisLP, that places labels based on the estimation of visibility. It employs machine-learning techniques to estimate the visibility that reflects human’s subjective mental workload in reading information and objective measures of reading correctness in various projection conditions. Li et al. \cite{ref58,ref59} presented empirical results extracted from experiments aimed at the user’s visual perception with regards to AR labeling layouts, reflecting the subjective preferences to different factors influencing the labeling result.

Unlike previous work, the approach presented in this paper considers not only the saliency information, but also the semantic information and the user labeling preferences in the label placement system. We also differ from previous task-unaware approaches by introducing a task-specific labeling framework and automatically learning the importance prior from the manually labeled dataset. 

\section{Our Proposed Method}
\label{sec:Metho}
\subsection{Overview}
\label{sec:Metho1}
In this section, we present our algorithm for semantic-aware task-specific label placement. Given a label, it needs to be placed in an appropriate position so as to avoid overlapping with other labels and salient regions from the user's view. A fundamental difference between previous works and our method is the fact that we consider the semantic information and specific task in addition to the classic model. These factors are able to contribute to the label placement system. We  introduce a new feature map called \textit{guidance map }to highlight the important regions in the user's view.
\par

\begin{figure*}[!htb]
  \includegraphics[width=1\textwidth]{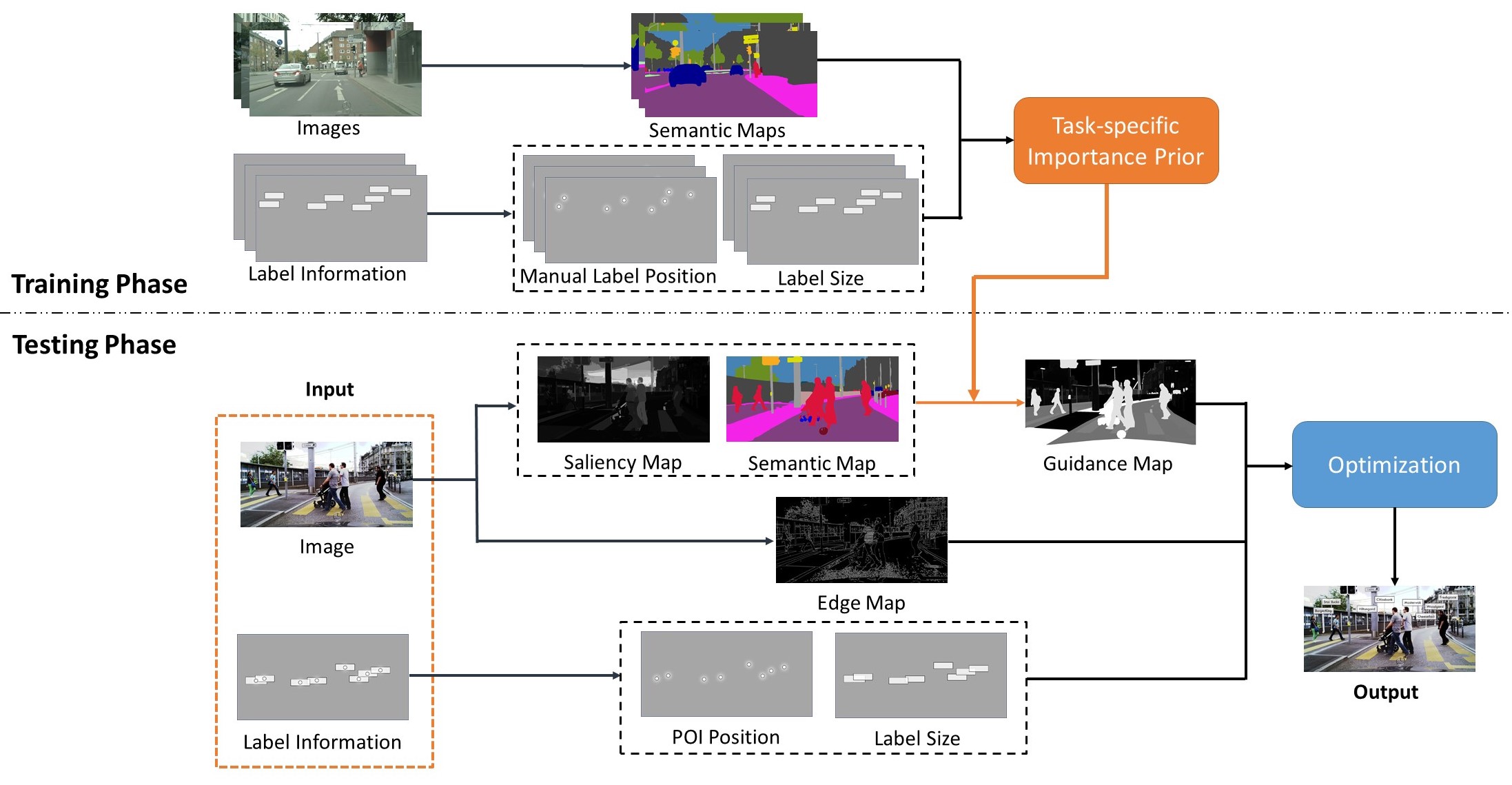}
\caption{Flowchart of the proposed algorithm}
\label{fig:1}     
\end{figure*}

Figure~\ref{fig:1} shows the processing flow of our method. During the training phase, we collect a \textit{Manual Label Placement }dataset and generate the task-specific importance prior from the semantic information and the manual labels. The importance prior  contributes to the saliency detection to get the guidance maps of the original images. At the testing phase, the label placement starts when the new object is detected in the field of projection corresponding to the preset labels or a new label placement is requested by the application. For an input image and preset label information, we first obtain the saliency map, semantic map and edge map, all from the original images. The input label information includes the POI position and the label size. Then we analyze the saliency map and semantic map with the task-specific importance prior learned from the training phase to generate the Guidance Map of the image. Finally, we add the guidance map with the label information into the optimization layout solver to output the user’s view image with the labels in their appropriate positions. 

\subsection{Creating the manual label placement dataset} \label{sec:Metho2}
To the best of our knowledge, there is no standard dataset for label placement in AR. Since we are using AR label placement in street view scenarios as a motivation application, we randomly select images from the Cityscapes Dataset \cite{ref46} to simulate the scenarios in the AR Street View navigation application. The reason for choosing the task of AR street view navigation is because urban scene images are widely used in auto-driving technology analysis. Afterwards, we collect manual label information through an experiment to generate the Manual Label Placement dataset. 

\textbf{Data selection and scene simulation}
In AR-based navigation, labels are used in the users' view to indicate the nearby restaurants, banks or other locations of interest. When collecting the dataset, the experiment participants are told to imagine the following task scenario: we are driving a car or walking on the road, and want to see additional information about the surrounding destinations of interest in an on-demand fashion. The AR street view navigation system connects the POI to the corresponding label with the help of a leader line, clearly associating the target location to the augmented overlaid information.

We randomly choose 300 pictures from the Cityscapes Dataset \cite{ref46} (180 for training and 120 for testing). The street scenes in this dataset are perfectly suited for the purpose of simulating a driving scenario. Moreover, the semantic annotation is very accurate such that the semantic information can be used in the ablation study. 

\begin{figure}[!htb]
  \includegraphics[width=0.5\textwidth]{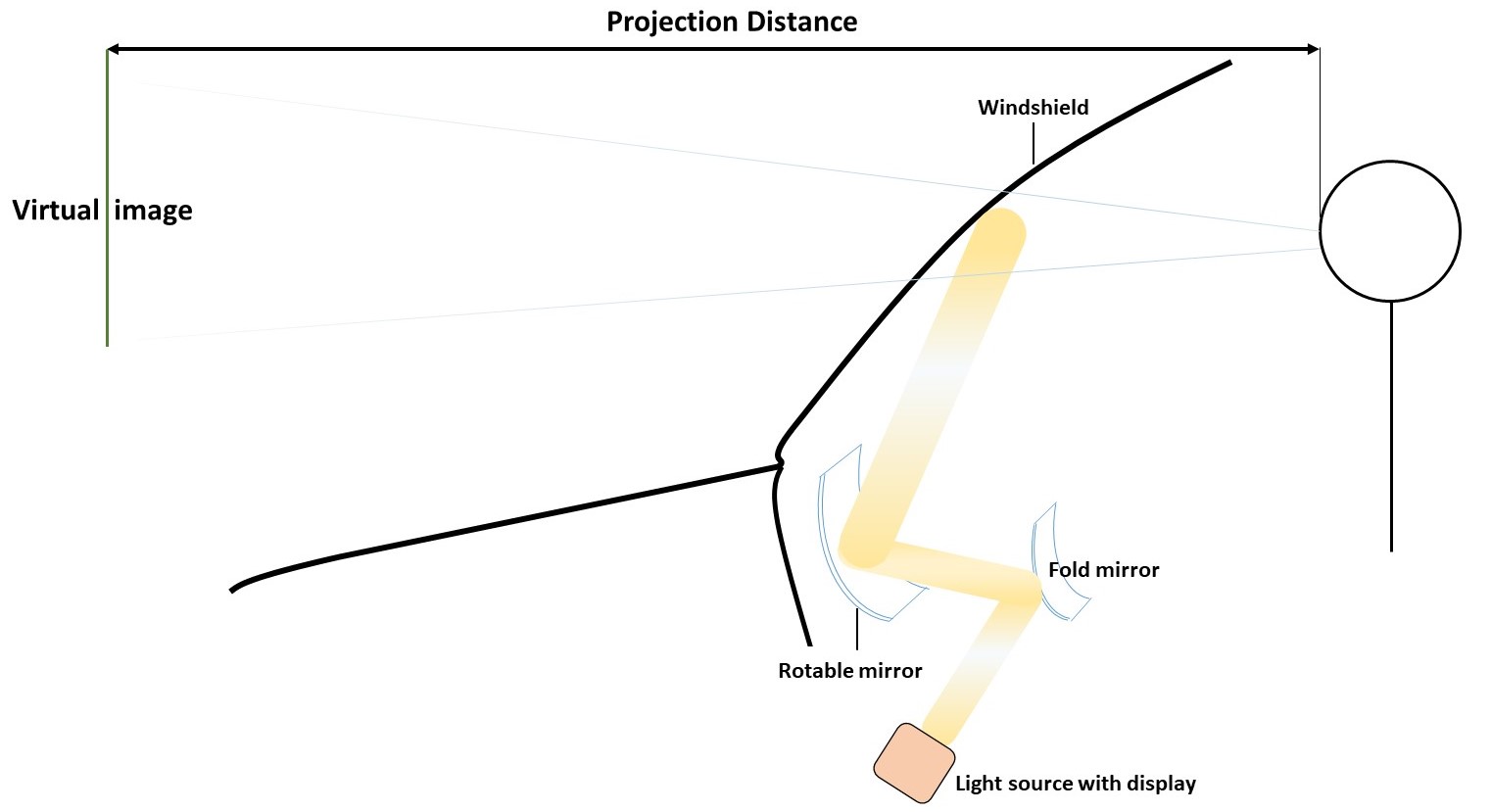}
\caption{Schematic diagram for AR navigation display in AR-HUD}
\label{fig:2}      
\end{figure}

We collect the manual placement data by simulating an AR street view navigation scenario to make it look as if the participants are sitting in a car and using equipment like AR-HUD. One of the most popular AR-HUD products produced by Continental inserts full-color graphics into the real road view in an approximately 130 cm-wide and over 60 cm-high section of the driver’s field of vision at a distance of 2.5 meters. The basis is provided by the digital micro-mirror device (DMD) technology, which is also used in digital cinema projectors, as shown in Figure~\ref{fig:2}.

\begin{figure*}[!htb]
  \includegraphics[width=17cm]{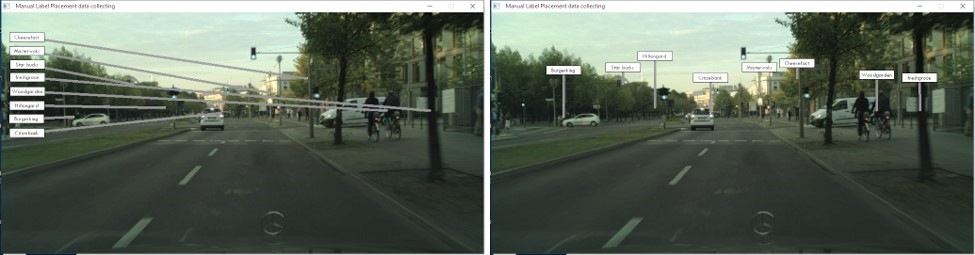}
\caption{Illustration for Manual Label Placement dataset collecting. Participants view the image with 8 labels, and use a mouse to drag the labels onto the appropriate place, just like the right image.}
\label{fig:3}     
\end{figure*}

Every participant is positioned in front of the projection screen at a distance of 2.5 meters. The valid section of the field of view is 120cm wide and 60cm high. The participants are asked to use a mouse to place labels onto the position where they deem most appropriate. 
\par

\textbf{Label configuration}
We need to set the POI point and its corresponding label information before the experiment. We should first consider the number of  target objects and related labels in the same street view image. If the number of labels in the user's view is low (1 or 2 labels), the layout method can easily avoid the important regions in the user's view, due to the abundance of candidate positions. In this case, the labels can be placed arbitrarily. If the number of labels in the image is high (more than 10), the label placement task can prove daunting even for humans, as it is very difficult to decide on an appropriate label layout. Therefore, we show the observers different images with the number of labels varying between 5 and 9, and also let them to vote for the maximum number of labels that they feel acceptable. Based on their selection, we made a suitable trade-off between visual quality and algorithm validity, and set the number of labels in each view to 8. As for POI point selection, we try to find meaningful regions like gates or windows as the target objects.
 
When placing labels, the content of labels may also affect the participant's decision. The AR annotation can be in various forms such as text, image, video, etc., and we set the content of the label as the name of the simulation target location. We noticed that different names may affect different user's labeling decision. In order to avoid users regarding some labels as more important than others in terms of the textual content of the labels, we opt to set the textual content of the 8 labels in each image with exactly the same names. The style of the label is black outline border, white background and black text.

The size of the labels may also affect the user's decision. We set the eight labels of the same size. We show examples of different label size styles and let the participants vote for their favorite. We finally set the text size to 12 pixels and the label size as $30\times120$ pixels.\par

\textbf{Collecting label placements}
We conduct the experiment to study how users manually place labels in specific scenes. There are 20 users (age range 18 to 40, 11 females and 9 males) participating in the study. The participants are asked to use a mouse to place labels onto the position they think most appropriate, just like Figure 3. In the experiment, all participants view the same 300 photos (180 for training and 120 for testing), and the order of the images is randomized for each participant. We finally collect $20\times300\times8=$ 48,000 labels.

\subsection{Guidance Map generation}
As we mentioned in Section 2, image-based AR label placement not only values the readability of the target objects and labels, but also uses the background and other important regions in the image of the user's view. When placing labels, previous work mostly uses the visual saliency to avoid important regions, however, the saliency detection methods in previous works are outdated. Therefore, we ran different saliency algorithm \cite{ref36,ref42,ref43} on the images in our dataset, the results are shown in Figure 4. 
\par

\begin{figure*}[!htb]
  \includegraphics[width=17cm]{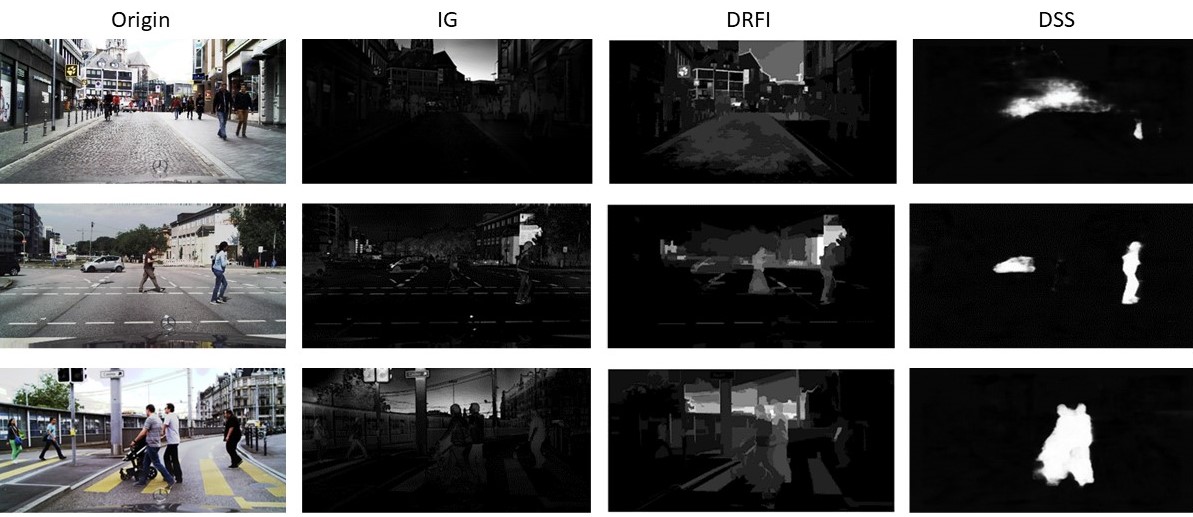}
\caption{Visual comparison of selected saliency detection.}
\label{fig:4}   
\end{figure*}

We found that the saliency detection methods do not perform well in street view images. For the contrast-based traditional methods like IG \cite{ref36} and DRFI \cite{ref42}, the sky regions in the results are shown to be more salient than the surroundings, which is contrary to our expectations. Individuals tend to place labels onto uniform regions, and the sky region is obviously on the top of the list of positions we expect users to place the labels on. Also, the road regions are less salient than others, and sometimes the traffic sign is not salient. This will lead the label placement algorithm to place the labels onto road or traffic sign regions, which will affect the drivers' view. So the saliency detection used in the previous works in street view image is not ideal. 
\par

\textbf{Task-specific importance prior}
Based on our observations, the semantic information in the scene has an influence on the user's tendency when placing labels. In a specific task scenario, the user's tendencies when placing labels can vary on based on the semantic regions. For this purpose, we integrate the semantic information with the saliency information into the label placement algorithm. But the question is, how to use the semantic information and how to measure the users' labeling preference on different semantic areas? We can treat the user's labeling preferences as a task-specific importance prior. In order to quantify the importance prior of the users' manual label placement in the AR Street View navigation scenario, we estimate the importance prior from human annotations using the collected data, specifically the 28,800 labels in the training set. We calculate the frequency of labels falling into regions of a specific semantic category.
\par

We also think it is important to not only look at the label's centroid for determining the semantic category that the label is placed in, but instead it is crucial to look at the accompanying category of each pixel belonging to a label. Just like the example shown in Figure 5, although the centroid of the label is related to the semantic category \emph{Foliage}, most of the label area is in the \emph{Sky} region. So we calculate the quota instead, as shown in Figure 5.
\par

\begin{figure}[!htb]
\centering
  \includegraphics[width=0.5\textwidth]{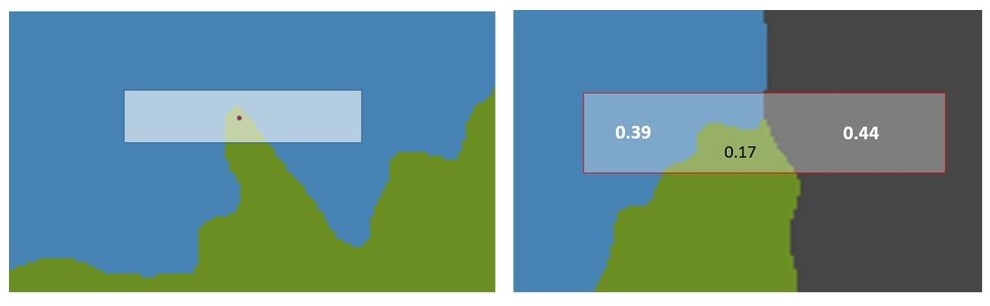}
\caption{An incorrectly calculated example (left) and the implemented method of calculation (right).}
\label{fig:5}  
\end{figure}

We calculate the number of labels for each semantic category in the training set, and name it as the number of actual labels $N_\mathrm{actual}$. However, this number can not directly represent the manual placement tendencies, because different semantic categories appear with different frequencies in the dataset. For example, most of the drivers think the \emph{Bridge} region is less important in their view and tend to put the labels onto it. The \emph{Bridge} semantic category only appears 9 times in our training set, and at most $9\times8\times20=$ 1,440 labels can be placed onto the \emph{Bridge} regions. We call 1,440 as the number of potential labels $N_\mathrm{potential}$ for the \emph{Bridge} semantic category. We introduce a tendency factor defined as:
\begin{equation} 
\lambda=\frac{N_\mathrm{actual}}{N_\mathrm{potential}}
\end{equation}

The ratio $\lambda$ for the \emph{Bridge} category is high enough, so even though the $N_\mathrm{actual}$ for the \emph{Bridge} category is very low, we can conclude that users tend to put the labels onto the \emph{bridge} region once they see it. The statistic placement tendencies for different semantic categories is shown in Table 1.
\par

\begin{table}[!htb]
\centering
\caption{Placement tendencies for different semantic categories}
\label{tab:1}      
\begin{tabular}{lccc}
\hline\noalign{\smallskip}
Categories & $N_\mathrm{potential}$ & $N_\mathrm{actual}$ & $\lambda$ \\
\noalign{\smallskip}\hline\noalign{\smallskip}
Sky & 28800 & 12569.43 & 43.64\% \\
Foliage & 28480 & 10840.36 & 38.06\% \\
Building & 28480 & 4956.41 & 17.40\% \\
Bridge & 1440 & 156.31 & 10.85\% \\
Person & 22720 & 97.52 & 0.43\% \\
Bus & 1120 & 3.04 & 0.27\% \\
Motorcycle & 1920 & 5.11 & 0.27\% \\
Pole & 28800 & 51.31 & 0.18\% \\
Sidewalk & 28800 & 47.21 & 0.16\% \\
Car & 27680 & 40.36 & 0.15\% \\
Traffic Sign & 27040 & 37.64 & 0.14\% \\
Rider & 5280 & 6.43 & 0.12\% \\
Road & 28000 & 15.76 & 0.05\% \\
Bicycle & 7840 & 2.28 & 0.03\% \\
Traffic Light & 13600 & 0.83 & 0.01\% \\
\noalign{\smallskip}\hline
\end{tabular}
\end{table}

From Table 1 and an interview that we performed with the participants after the data collection, we can confirm that the statistic manual label placement tendencies are reasonable. As shown, \emph{Sky}, \emph{Foliage}, \emph{Building} and \emph{Bridge} are among the most widely chosen categories to put the label on. The \emph{Sky} and \emph{Foliage} region is very common in the dataset, and drivers tend to put labels on these regions since the color and texture are uniform and don't vary drastically. Sometimes these regions are more salient than surroundings in the saliency map, leading algorithms to avoid placing labels in these regions. Also, even if the \emph{Road} regions constitute a great proportion in the dataset, drivers do not like putting the labels on it since they think the labels will adversely effect their driving. In spite of the \emph{Traffic Sign} and \emph{Traffic Light}'s appearance in almost every picture in the training dataset and their high $N_\mathrm{potential}$ value, drivers do not put labels on it. We conclude that the reason for this is because they think these are the most important object they want to see while driving, and will not allow anything to cover it.
\par

\textbf{Guidance Map}
We use the semantic information and the preference of user's label placement to adjust the saliency map to generate a Guidance Map, closely resembling users' natural preferences to indicate the important regions in the view in a specific task. The Guidance Map, denoted by $G$, can be conducted through the model:
\begin{equation}
G(i,j)=\frac{H^\prime(i,j)\cdot{S(i,j)}}{\max(S(i,j))}
\end{equation}
\begin{equation}
H^\prime(i,j)=c_{ij}\cdot{H(i,j)}
\end{equation}
where, for pixel location $(i,j)$, $S(i,j)$ represents the saliency map, $H(i,j)$ represents the semantic map, $\max(\cdot)$ is the maximum function for normalization, and $c$ is the task-specific importance prior weight learned in the training phase. In this study, from the statistic features for manual placement tendency, we define $c$ for different semantic regions as the following
\begin{equation}
c=1-\frac{\lambda}{{\left\lVert \lambda\right\lVert}_{\infty}}
\end{equation}
Table 2 shows the task-specific category weights in this study. Even though some participants put some labels on the \emph{traffic sign} and \emph{traffic light} regions, we assume that the participants did not notice these important objects due to not paying enough attention. We conclude that this is mainly because the users expressed their regret with regards to these labels during our interview. For this purpose, we adjusted these coefficients to 1.
\begin{table}[!htb]
\centering
\caption{Task-specific importance prior weight for different semantic categories}
\label{tab:2}   
\begin{tabular}{lcc}
\hline\noalign{\smallskip}
Categories & $\lambda$ & $c$ \\
\noalign{\smallskip}\hline\noalign{\smallskip}
Sky & 43.64\% & 0.0000 \\
Foliage & 38.06\% & 0.1279\\
Building & 17.40\% & 0.6013\\
Bridge & 10.85\% & 0.7514\\
Person & 0.43\% & 0.9901\\
Bus & 0.27\% & 0.9938\\
Motorcycle & 0.27\% & 0.9938\\
Pole & 0.18\% & 0.9959\\
Sidewalk & 0.16\% & 0.9963\\
Car & 0.15\% & 0.9966\\
Traffic Sign & 0.14\% & 1.0000\\
Rider & 0.12\% & 0.9973\\
Road & 0.05\% & 0.9989\\
Bicycle & 0.03\% & 0.9993\\
Traffic Light & 0.01\% & 1.0000\\
\noalign{\smallskip}\hline
\end{tabular}
\end{table}
\par
\begin{figure*}[!htb]
\centering
  \includegraphics[width=17cm]{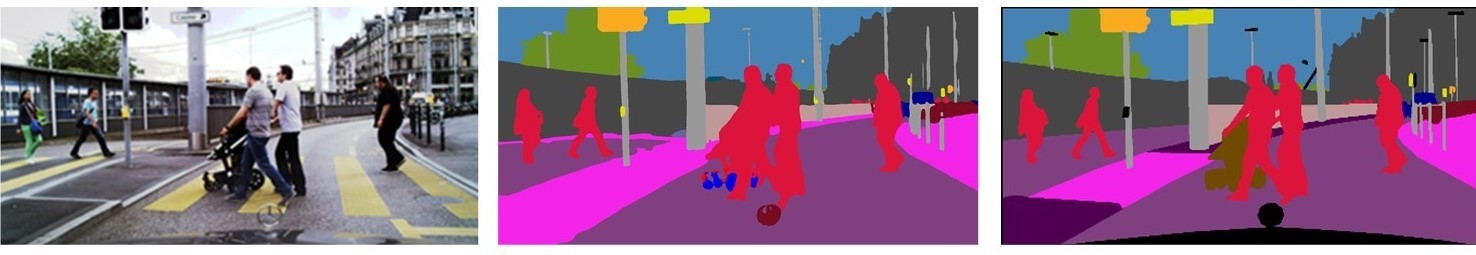}
\caption{Qualitative result of semantic segmentation generated by Deeplabv3 (middle), the origin image (left) and ground-truth (right).}
\label{fig:6}       
\end{figure*}
\begin{figure*}[!htb]
\centering
  \includegraphics[width=17cm]{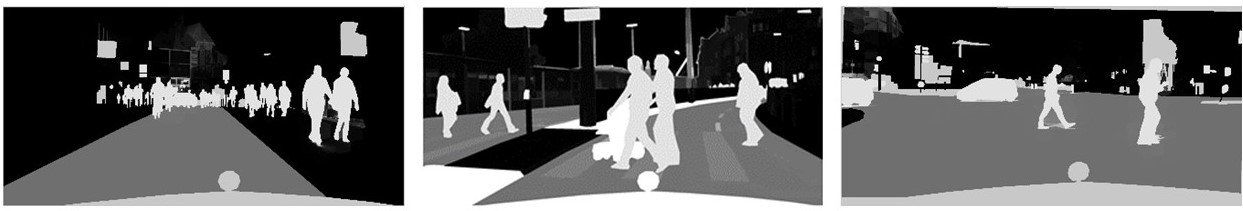}
\caption{Qualitative result of Guidance Map generated by the proposed method.}
\label{fig:7}       
\end{figure*}
To obtain the important regions in the user's view, we combine the saliency-detection algorithm with semantic information for the image of the user’s view. The pixel’s value indicates the importance of the region for the drivers. Also, we can see edges do not always show as significantly salient after running the saliency-detection algorithm. It is obvious that edges of objects should not be occluded by labels. For this purpose, we use the Canny algorithm to detect the edges of objects in the user's view, serving as a supplement for the Guidance Map, obtaining the positions where annotations tend not to be placed. 
\par

In our system, we use the DSS saliency-detection algorithm proposed by Cheng \cite{ref43}. The semantic information comes from semantic segmentation result of Deeplabv3 \cite{ref51}, which shows good performance in the City-scapes dataset, as shown in Figure 6. Our Guidance Map is shown in Figure 7.
\par

\subsection{Optimization}
\label{sec:Metho4}
Throughout the paper, for the $t$-th image in the testing set ($T$ images), we let the target objects in the $t$-th image form a set $O=\left\{O_{1}, \ldots, O_{k}, \ldots, O_{K}\right\}$ , $O_{k}$ represent the $k$-th target object, and $K$ is the total number of objects. Then we set the POI as the centroid of an object's bounding box and named as $\vec{m}_{k}=(x_{k},y_{k})$. To present the visual information in our study, each $O_{k}$ has its own label $P_{k}$, and the position of the label $P_{k}$ is $\vec{p}_{k}=(\tilde{x}_{k},\tilde{y}_{k})$. The set $P=\left\{P_{1}, \ldots, P_{k}, \ldots, P_{K}\right\}$ is defined as the set of labels in the $t$-th image. Then the label placement problem aims to compute each $\vec{p}_{k}$. 
\par
We formulate the label placement problem as an energy minimization problem, with the energy function defined as
\begin{equation}
\begin{split}
\mathcal{E}=\mathcal{E}_{\mathrm{lb}}+\mathcal{E}_{\mathrm{ln}}
\end{split}
\end{equation}
\par
It consists of the label energy term ($\mathcal{E}_{\mathrm{lb}}$) and the leader line energy term ($\mathcal{E}_{\mathrm{ln}}$).

\textbf{Label energy term}
As shown in Eq. 6, the label energy term includes the energy of the label overlaying the Guidance Map and edge map, and the energy of the labels overlaying each other.
\begin{equation}
\begin{split}
\mathcal{E}_{\mathrm{lb}}=\alpha_{1} \mathcal{E}_{\text{lb-g}}+\alpha_{2} \mathcal{E}_{\mathrm{lb\text{-}e}}+\alpha_{3} \mathcal{E}_{\mathrm{lb\text{-}int}}
\end{split}
\end{equation}
where $\alpha_{i} (i=1,2,3)$ are the weight coefficients, which are \textit{automatically} learned from the Manual Label Placement dataset.
\begin{enumerate}[(1)]
\item Label overlaying Guidance Map (joint saliency \& semantic)
\begin{equation}
\mathcal{E}_{\mathrm{lb\text{-}g}}= \sum_{k=1}^{K} \sum_{i=1}^{\mu} \sum_{j=1}^{\nu}  \frac{M_{P_k}(i,j) \cdot G(i,j)}{\mu_{L} \cdot \nu_{L}}
\end{equation}
where the $M_{P_k}(i,j)$ is the map which indicates the region of the annotation of $P_k$ (the label area pixel value is $1$ and the rest of the image is $0$), the image size is $\mu \cdot \nu$ and the label size is $\mu_{L} \cdot \nu_{L}$.
\par

\item Label overlaying the edge map (Canny)
\begin{equation}
\mathcal{E}_{\mathrm{lb\text{-}e}}= \sum_{k=1}^{K} \sum_{i=1}^{\mu} \sum_{j=1}^{\nu}  \frac{M_{P_k}(i,j) \cdot E(i,j)}{\mu_{L} \cdot \nu_{L}}
\end{equation}
where the $E(i,j)$ is the edge map generated by Canny edge detector. 

\item Labels intersection
\begin{equation}
\mathcal{E}_{\mathrm{lb\text{-}int}}= \sum_{P_k,P_{k^\prime} \in P} \sum_{i=1}^{\mu} \sum_{j=1}^{\nu}  \frac{M_{P_k}(i,j) \cdot M_{P_{k^\prime}}(i,j)}{\mu_{L} \cdot \nu_{L}}
\end{equation}

\end{enumerate}

\textbf{Leader line energy term}
As shown in Eq. 10, the label energy term includes the energy of the leader line overlaying the Guidance Map, the leader line intersection, the leader line length and orientation.
\begin{equation}
\begin{split}
\mathcal{E}_{\mathrm{ln}}=\beta_{1} \mathcal{E}_{\mathrm{ln\text{-}g}}+\beta_{2} \mathcal{E}_{\mathrm{ln\text{-}int}}+\beta_{3} \mathcal{E}_{\mathrm{ln\text{-}len}}+\beta_{4} \mathcal{E}_{\mathrm{ln\text{-}ori}}
\end{split}
\end{equation}
where $\beta_{i} (i=1,\ldots,4)$ are the weight coefficients, which are automatically learned from the Manual Label Placement dataset.

\begin{enumerate}[(1)]

\item Leader line overlaying Guidance Map
\begin{equation}
\mathcal{E}_{\mathrm{ln\text{-}g}}= \sum_{k=1}^{K} \sum_{i=1}^{\mu} \sum_{j=1}^{\nu}  \frac{N_{P_k}(i,j) \cdot G(i,j)}{\mu_{L} \cdot \nu_{L}}
\end{equation}
where the $N_{P_k}(i,j)$ is the map which indicate the region of the leader line between the label $P_k$ and the POI (the leader line pixel value is $1$ and the rest of the image is $0$).

\item Leader line intersection
\begin{equation}
\mathcal{E}_{\mathrm{ln\text{-}int}}= \sum_{P_k,P_{k^\prime} \in P} \sum_{i=1}^{\mu} \sum_{j=1}^{\nu}  {N_{P_k}(i,j) \cdot N_{P_{k^\prime}}(i,j)}
\end{equation}

\item Leader line length
\begin{equation}
\mathcal{E}_{\mathrm{ln\text{-}len}}=\sum_{k=1}^{K}  {\left|\vec{p}_{k}-\vec{m}_{k}\right|}
\end{equation}

\item Leader line orientation
\begin{equation}
\mathcal{E}_{\mathrm{ln\text{-}ori}}=\sum_{k=1}^{K}  {\left|\cos(\phi(\vec{p}_{k}-\vec{m}_{k}))\right|}
\end{equation}
where the $\phi(\vec{p}_{k}-\vec{m}_{k})$ is the angle between the leader line vector $\vec{p}_{k}-\vec{m}_{k}$ and the $y$ axis (the vertical alignment leader line is preferred). 
\end{enumerate}
\par

We consider three algorithms for implementing the optimization: greedy algorithm, simulated annealing and a force-based algorithm. We firstly rule out simulated annealing algorithm because its low efficiency is not suitable for our particular AR scenario. The greedy algorithm iterates all label and calculates their energy function values. We find the minimum of these values to determine the final appropriate positions. The force-based algorithm implements penalty factors as a set of forces, and labels are moved in parallel in this force field. After a certain number of iterations, we get the labels' final positions. When testing the force-based algorithm, we found that the force-field is too complex for us to handle. It is impossible to find the appropriate number of iterations in the implementation. Therefore, we chose the greedy algorithm for our optimization. We sort the labels from left to the right, nearest to the farthest. Then we iterate each label for calculation. In the end, we find the minimum and obtain the final results.
\par

\section{Experiments and Results}
\label{sec:exper}
We apply our proposed method to the Manual Label Placement dataset, which has 180 images for training and 120 for testing. We use the training set to generate the task-specific semantic-aware weights and the coefficients in the energy function as $\alpha_{1}=0.3514$, $\alpha_{2}=0.0675$, $\alpha_{3}=0.0839$ and $\beta_{1}=0.0371$, $\beta_{2}=0.1078$, $\beta_{3}=0.2302$, $\beta_{4}=0.1221$. Afterwards, we generate the label layout on the testing dataset. Qualitative and quantitative comparison is conducted between the proposed method and state-of-the-art label placement algorithms.

\subsection{Evaluation Metrics}
\label{sec:exper1}
To evaluate the performance of the annotation results, we apply four metrics, (1) the average distance from the centroid of manual placement $\mu_{\mathrm{centroid}}$, (2) the average annotation overlapped area $\mu_{\mathrm{over}}$, (3) the intersection $\mu_{\mathrm{int}}$, and (4) the leader line length $\mu_{\mathrm{len}}$.
\par
The $\mu_{\mathrm{centroid}}$ metric is one of the most important metrics. It aims to assess the difference between the implementation result and users’ manual placement. In the testing dataset, 20 participants decide the label positions for each label $P_{k} \in P$ for $T$ images. We eliminate two of the most isolated and get the centroid of the remaining 18 positions to set it as $\tilde{\vec{p}}_{k}$. The $\mu_{\mathrm{centroid}}$ is defined as
\begin{equation}
\mu_{\mathrm{centroid}}=\frac{1}{K \cdot T} \sum_{t=1}^{T} \sum_{k=1}^{K} \left|\vec{p}_{k}^{t}-\tilde{\vec{p}}_{k}^{t}\right|
\end{equation}
A lower value of $\mu_{\mathrm{centroid}}$ means that the placement result is closer to the manual placement.
\par

The second metric $\mu_{\mathrm{over}}$ aims to assess the severity of overlapping of both annotations and important regions. $\mu_{\mathrm{over}}$ is defined as:
\begin{equation}
\mu_{\mathrm{over}}=\frac{1}{K \cdot T} \sum_{t=1}^{T} ( \mathcal{E}_{\mathrm{lb-g}}^{t} + \mathcal{E}_{\mathrm{lb-int}}^{t} )
\end{equation}
A lower value of $\mu_{\mathrm{over}}$ means that fewer collisions occur between annotations and important regions. 
\par

The third metric $\mu_{\mathrm{int}}$ aims to estimate if label leader lines have intersections. To prevent confusion when reading the annotations, the leader lines should not intersect with each other. To assess the severity of intersection, $\mu_{\mathrm{int}}$ is represented as:
\begin{equation}
\mu_{\mathrm{int}}=\frac{1}{T} \sum_{t=1}^{T} \mathcal{E}_{\mathrm{ln-int}}^{t}
\end{equation}
Fewer intersections of leader lines yield to a lower $\mu_{\mathrm{int}}$.
\par

The last evaluation metric is evaluating the leader line length. If the label is not close to the point of interest, users need to spend time to track the related label. Moreover, the participants stated in the interview that they dislike long leader lines and think that the acceptable leader line length is 5-10 cm . We set $\gamma$ as the optimized length, $\mu_{\mathrm{len}}$ is computed to quantify the similarity between $|\vec{p}_{k}^{t}-\vec{m}_{k}^{t}|$ and $\gamma$ as: 
\begin{equation}
\mu_{\mathrm{len}}=\frac{1}{T} \sum_{t=1}^{T} \sum_{k=1}^{K} \left(\left|\vec{p}_{k}^{t}-\vec{m}_{k}^{t}\right|-\gamma\right)
\end{equation}
When the leader line length is closer to the optimal length, the value of $\mu_{\mathrm{len}}$ will be smaller. We set $\gamma=10$ in our experiment.
\par
\subsection{Qualitative comparison with state of the art}
\label{sec:exper2}

\begin{figure*}[!htb]
\centering
  \includegraphics[width=17cm]{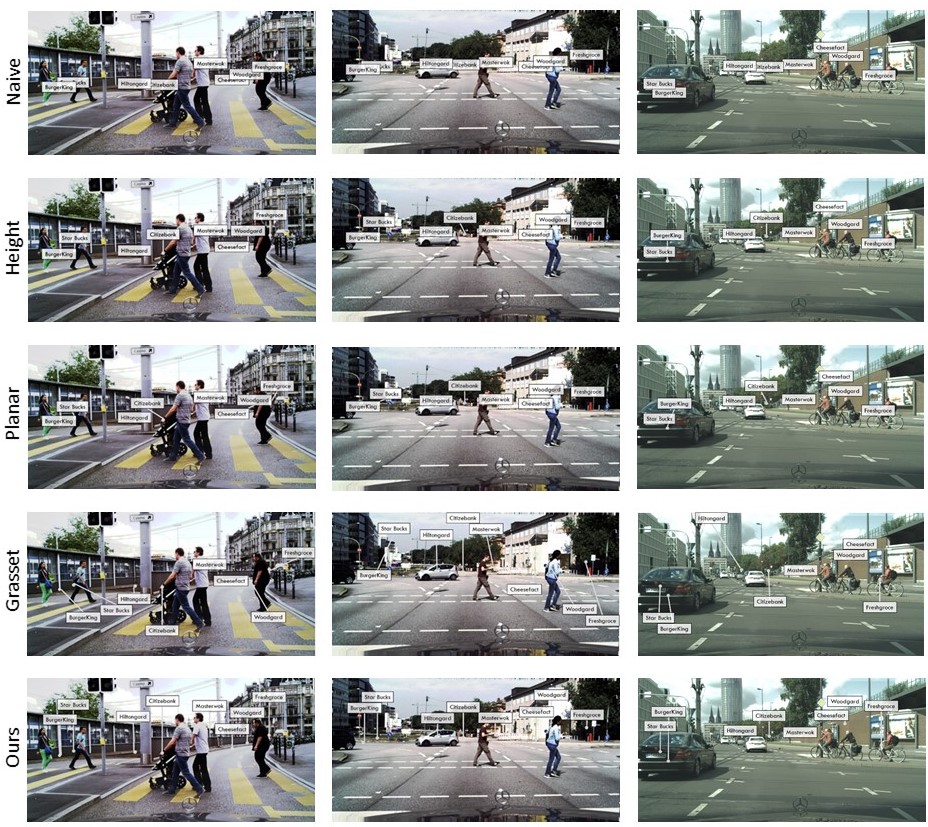}
\caption{The qualitative comparison between the proposed method and other state-of-the-art methods.}
\label{fig:8}      
\end{figure*}
The comparisons with the state-of-the-art approaches help identify the contributions of the proposed approach. We compare the proposed approach to a baseline approach and three state-of-the-art approaches: height-separation \cite{ref55}, planar-separation and Grasset's method~\cite{ref17}. The baseline approach is the naive method which overlays the computer-generated information onto the POIs in the user’s view. Naturally, this approach leads to label occlusions. In the height separation method presented by Peterson et al. \cite{ref55}, the annotation and the POI share the same $x$-coordinate. The height separation technique iterates each label, and when it detects that two labels overlap, it moves up the label related to the farthest POI for half a label's height. Similar to the naive and the height-separation method, the planar separation technique also lacks image analysis. With planar separation, if a label is detected to overlap another label, it will evaluate the new label position with $10^{\circ}$ separations and choose the best position after evaluating a total of 36 positions. Additionally, we also compare our technique to the layout algorithm presented by Grasset \cite{ref17}. Qualitative results are shown in Figure 8. 
\par
From the qualitative comparison, we can see that the naive approach, the height separation method and the planar separation method inappropriately cover pedestrians and traffic signs. Moreover, there is a significant inter-overlapping of labels with the naive method, which seriously affects the readability of the labels. This case necessitates that the drivers change the viewpoint to see the annotations. Grasset's \cite{ref17} method is based on the IG saliency algorithm to avoid occluding pedestrians and traffic signs, and a large number of the labels are placed on the road, which strongly affect the driver's vision in the AR street view navigation scene.
\par
The above mentioned problems are eliminated in our method. The labels do not cover important areas in the user's field of view, and the labels tend to be positioned on regions of uniform color and texture (sky, leaves, bridge), while avoiding the road as expected. Our method outperforms other methods on our testing dataset.

\subsection{Quantitative Results and Ablation Study}
\label{sec:exper3}
From the evaluation metrics defined in Section 4.1, we quantitatively compare our method with the state-of-the-art approaches. The quantitative comparison is shown in Table 3.
\par

\begin{table}[!htb]
\caption{Quantitative Results of Different Methods}
\label{tab:3}       
\begin{tabular}{lcccc}
\hline\noalign{\smallskip}
Methods & $\mu_{\mathrm{centroid}}$ & $\mu_{\mathrm{over}}$ & $\mu_{\mathrm{int}}$ & $\mu_{\mathrm{len}}$\\
\noalign{\smallskip}\hline\noalign{\smallskip}
Naive & 105.16 & 38.74 & 0 & 0 \\
Height Separation & 86.39 & 83.30 & 0 & 10.73 \\
Planar Separation & 89.45 & 48.88 & 0 & 9.84 \\
Grasset\cite{ref17} & 85.33 & 35.58 & 78 & 123.43 \\
Proposed Method & 61.78 & 8.50 & 12 & 26.99 \\
\noalign{\smallskip}\hline
\end{tabular}
\end{table}

From the table, we can see that for one of the most important evaluation metric $\mu_{\mathrm{centroid}}$, our method has the lowest value compared to others - our label layout is closer to the Manual Label Placement benchmark and hence provides better layout than the comparing methods. For another important metric $\mu_{\mathrm{over}}$, out method yields a much smaller value than other methods, indicating that we successfully avoid occluding the salient areas in the user's view. As for $\mu_{\mathrm{int}}$ and $\mu_{\mathrm{len}}$, it is obvious that the first three methods will get lower values. Out method performs much better than \cite{ref17} based on the two metrics. To summarize, our method performs better than the other state of the art methods.
\par

To validate the impact of the different components of the latest saliency model and the semantic information, we conduct ablation experiments on the testing dataset, shown in Table 4.
\par

\begin{table}[!htb]
\caption{Ablation Study}
\label{tab:4}       
\begin{tabular}{lcccc}
\hline\noalign{\smallskip}
Methods & $\mu_{\mathrm{centroid}}$ & $\mu_{\mathrm{over}}$ & $\mu_{\mathrm{int}}$ & $\mu_{\mathrm{len}}$\\
\noalign{\smallskip}\hline\noalign{\smallskip}
IG(Grasset\cite{ref17}) & 85.33 & 35.58 & 78 & 123.43 \\
IG+Deeplabv3 & 72.70 & 15.18 & 27 & 46.83 \\
IG+Groundtruth & 64.65 & 9.17 & 16 & 37.48 \\
DSS+Deeplabv3(Ours) & 61.78 & 8.50 & 12 & 26.99 \\
DSS+Groundtruth & 57.65 & 7.16 & 12 & 26.49 \\
\noalign{\smallskip}\hline
\end{tabular}
\end{table}

The ablation study results indicate that adding the semantic information to the IG algorithm, the label layout quantitative evaluation performs better. When directly integrating the accurate semantic segmentation ground truth, the improvement is more obvious. It indicates that the semantic information is useful for the label placement problem. On the other hand, from the comparison of DSS+Deeplabv3 (our method) with IG+Deeplabv3, we can conclude that the latest state-of-the-art saliency model improves the label layout performance. These studies show that each ingredient brings individual improvement, and all of them work together to produce better label layouts.
\par

\section{Conclusion}
\label{sec:concl}
This paper presents a semantic-aware approach for label placement in AR street view scenarios. We introduce a new algorithm for labeling that can be used for future development of AR-HUD street view navigation applications. Compared to other label layout algorithms, our method has the following advantages: (1) Both semantic information and saliency detection are integrated into the label placement optimization to further improve the layout performance. (2) With the help of a label placement dataset, we have a quantitative evaluation benchmark to conduct the quantitative experiment. (3) Unlike previous task-unaware methods, our system provides a task-specific label placement framework. 
\par

\end{document}